%% file: gem_occ_arxiv.tex
\documentclass{article}
\usepackage{iclr2027_conference,times}
\input{arxiv_preamble}

\title{From Visual Geometry Evidence to Embodied Semantic Occupancy Memory}
\newcommand{\gemaffmark}[1]{\raisebox{0.6ex}{\fontsize{7}{7}\selectfont #1}}
\author{%
Hu Zhu\gemaffmark{1,5}\quad
Bohan Li\gemaffmark{2,5}\quad
Xianda Guo\gemaffmark{3}\quad
Yanlun Peng\gemaffmark{4}\quad
Hongsi Liu\gemaffmark{5}\quad
Baorui Peng\gemaffmark{6}\\[3pt]
Xiaofeng Wang\gemaffmark{7}\quad
Mingqi Yuan\gemaffmark{8}\quad
Xin Jin\gemaffmark{5}\quad
Wenjun Zeng\gemaffmark{5}\quad
Chang Wen Chen\gemaffmark{1}%
}
\newcommand{\gemaffiliations}{%
\textsuperscript{1}The Hong Kong Polytechnic University\quad
\textsuperscript{2}Shanghai Jiao Tong University\\
\textsuperscript{3}Wuhan University\quad
\textsuperscript{4}Great Wall Motor\quad
\textsuperscript{5}Eastern Institute of Technology, Ningbo\\
\textsuperscript{6}Georgia Institute of Technology\quad
\textsuperscript{7}Tsinghua University\quad
\textsuperscript{8}The University of Hong Kong%
}
\newcommand{\gemresourcelinks}{%
\href{https://zhuhu00.top/GEM-Occ/}{\textcolor{gemresourceblue}{\faGlobe\enspace Page}}\qquad
\href{https://github.com/zhuhu00/GEM-Occ}{\textcolor{gemresourceblue}{\faGithub\enspace Code}}\qquad
\href{https://huggingface.co/datasets/zhuhu00/HI-Occ}{\textcolor{gemresourceblue}{\faDatabase\enspace Data}}%
}
\hypersetup{%
  pdftitle={From Visual Geometry Evidence to Embodied Semantic Occupancy Memory},
  pdfauthor={Hu Zhu, Bohan Li, Xianda Guo, Yanlun Peng, Hongsi Liu, Baorui Peng, Xiaofeng Wang, Mingqi Yuan, Xin Jin, Wenjun Zeng, Chang Wen Chen}%
}
\makeatletter
\renewcommand{\@maketitle}{%
  \vbox{\hsize\textwidth\centering
    {\fontsize{17}{20}\selectfont\bfseries\@title\par}
    \vskip 12pt
    {\fontsize{11}{14}\selectfont\@author\par}
    \vskip 7pt
    {\fontsize{9}{11.5}\selectfont\gemaffiliations\par}
    \vskip 9pt
    {\hypersetup{hidelinks}\fontsize{10}{12}\selectfont\gemresourcelinks\par}
    \vskip 8pt
  }%
}
\makeatother
\iclrfinalcopy
\begin{document}
\fancyhead{}
\renewcommand{\headrulewidth}{0pt}
\maketitle

\begin{abstract}
\input{arxiv_abstract}
\end{abstract}

\section{Introduction}

Embodied semantic occupancy mapping requires an agent to build a persistent spatial representation from observations acquired during exploration.
Semantic occupancy provides a natural basis for this representation by describing occupied structures and their semantic categories~\citep{song2017sscnet,cao2022monoscene,wu2025embodiedocc}.
As the agent moves between rooms, previously observed structures leave the current view but remain relevant to its understanding of the environment.
Returning to these regions provides new evidence that may reinforce earlier estimates or reveal errors caused by incomplete views and uncertain predictions.
A useful occupancy memory must therefore retain spatial knowledge beyond the current observation while remaining open to revision.
This requires distinguishing occupied structures, observed free space, and unknown regions, and maintaining these distinctions as observations accumulate across connected indoor spaces.

\begin{figure}[t]
    \centering
    \includegraphics[width=0.85\linewidth]{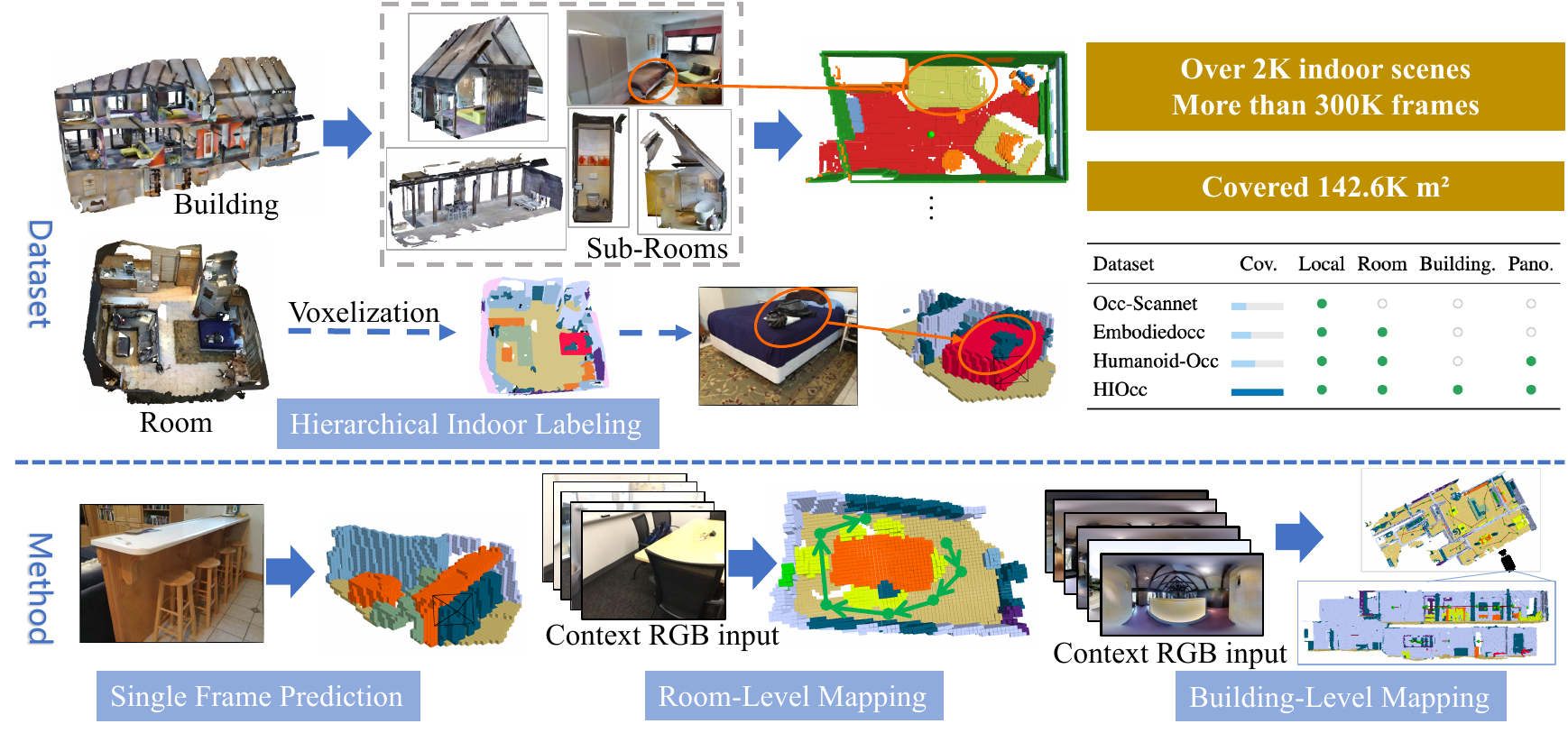}
    \caption{
    \textbf{Overview of \datasetname and \methodname.}
    \datasetname provides hierarchical indoor semantic occupancy annotations across local views, rooms, and connected buildings. 
    \methodname builds persistent semantic Gaussian memory for local prediction, room-level online mapping, and building-level mapping.
    }
    \label{fig:teaser}
\end{figure}

Recent voxel-based and Gaussian-based methods have improved local semantic occupancy prediction~\citep{cao2022monoscene,yu2024iso,qian2026splatssc,zhou2026gpocc}, while online methods extend perception through persistent scene representations.
EmbodiedOcc~\citep{wu2025embodiedocc} maintains a global Gaussian memory and progressively refines regions within the current view.
Its uniform initialization across a predefined scene extent, however, allocates memory to regions before they are observed, with the initial cost growing with scene volume.
Incremental Gaussian fusion provides another route to constructing global maps from successive predictions~\citep{zhou2026gpocc}.
For continued exploration, memory construction must address both where new elements are allocated and how accumulated estimates are revised.
Supporting observations should consolidate an existing estimate, observed free space should suppress contradicted occupancy, and occluded regions should retain their state in the absence of new evidence.
These cases can occur at the same location at different times, making observation confidence and visibility central to memory updates.
As the explored area grows, the accumulated representation must also support efficient access for subsequent updates and occupancy queries.

We introduce \methodname, a Gaussian Evidence Memory framework in which incoming observations determine both memory allocation and state updates.
For each posed observation, predicted geometry and semantics are converted into occupied semantic Gaussians and free-space ray evidence in a shared world coordinate system.
Geometry and uncertainty determine the spatial support of the Gaussians, while visible ray segments identify space observed to be empty.
Incoming Gaussians are associated with existing memory elements or inserted to represent newly observed structure.
Confidence-weighted fusion consolidates geometric and semantic estimates, and free-space evidence revises contradicted occupancy without treating occlusion as evidence of absence.
Spatial submaps and a building-level connectivity graph organize the growing memory, while merging and pruning consolidate redundant or poorly supported elements.
The resulting representation supports occupancy queries throughout exploration.
Training supervises both local predictions and fused memory outputs, connecting observation-level evidence construction with the quality of the accumulated map.

Evaluating this capability requires measuring how map quality, consistency, and computational cost evolve as observations accumulate.
We introduce \datasetname, a unified benchmark constructed from ScanNet, ScanNet++, and Matterport3D~\citep{scannet,scannetpp,Matterport3D2017}.
As summarized in Fig.~\ref{fig:teaser} and Table~\ref{tab:datasets}, it comprises over 2,000 scenes and approximately 300,000 observations, covering about 142,600\,m$^2$.
The benchmark combines perspective and panoramic observations under a shared semantic label space, sparse occupancy format, and evaluation framework.
Its three tracks connect local prediction, room-level online mapping, and building-level exploration, measuring the quality of individual predictions, their integration over time, and memory construction across connected rooms.
Occupancy accuracy and progress AUC measure map quality over exploration; revisit consistency measures agreement when previously observed regions are revisited; and memory footprint and query latency quantify storage and access costs.
Together, these evaluations connect the quality of local evidence with the reliability and efficiency of persistent occupancy memory.

Our contributions are summarized as follows:
\begin{itemize}
    \item We propose \methodname, a Gaussian Evidence Memory framework that allocates memory incrementally and consolidates occupied semantic and free-space evidence through confidence- and visibility-aware causal updates. 

    \item We introduce \datasetname, a unified indoor semantic occupancy benchmark spanning local prediction, room-level online mapping, and building-level exploration. 

    \item Experiments demonstrate improved semantic occupancy accuracy and revisit consistency, together with reduced memory consumption and query latency relative to flat Gaussian memory in building-level mapping.
\end{itemize}

\begin{table*}[t]
\centering
\caption{
\textbf{Occupancy dataset comparison.}
\textbf{Mod.}: C/L/R/D = camera/LiDAR/radar/depth. 
\textbf{Coverage Area}: approximate route or scene footprint.
For \datasetname, the 12 labels comprise 11 occupied semantic classes and one free-space label.
}
\renewcommand\tabcolsep{3.2pt}
\resizebox{1.0\textwidth}{!}{
\begin{tabular}{l|c|c|c|c|c|c|c}
\toprule
\textbf{Dataset} 
& \textbf{Source Data} 
& \textbf{Mod.} 
& \textbf{Surround Views} 
& \textbf{Scenes} 
& \textbf{Frames} 
& \textbf{Classes}  
& \textbf{Coverage Area} \\
\midrule

\rowcolor[HTML]{EFEFEF} 
\multicolumn{8}{c}{\textbf{Outdoor / Autonomous Driving}} \\
\midrule

SemanticKITTI~\citep{behley2019semantickitti} 
& SemanticKITTI, KITTI
& C+L 
& \textcolor{red}{\usym{2717}} 
& 22 
& 4.6K 
& 19 
& $\sim$3.9 km$^2$ \\

SSCBench~\citep{li2024sscbench} 
& KITTI-360, nuScenes, Waymo 
& C+L 
& \textcolor{ForestGreen}{\usym{2713}} 
& 1859 
& 67K 
& 19 
& $\sim$43.5 km$^2$ \\

SurroundOcc~\citep{wei2023surroundocc} 
& nuScenes 
& C+L 
& \textcolor{ForestGreen}{\usym{2713}} 
& 850 
& 40K 
& 17 
& $\sim$13.2 km$^2$ \\

OpenOccupancy~\citep{wang2023openoccupancy} 
& nuScenes 
& C+L 
& \textcolor{ForestGreen}{\usym{2713}} 
& 850 
& 34K 
& 17 
& $\sim$13.2 km$^2$ \\

OpenOcc~\citep{sima2023_occnet} 
& nuScenes 
& C+L 
& \textcolor{ForestGreen}{\usym{2713}} 
& 850 
& 40K 
& 17 
& $\sim$13.2 km$^2$ \\

Occ3D~\citep{tian2023occ3d} 
& nuScenes, Waymo
& C+L 
& \textcolor{ForestGreen}{\usym{2713}} 
& 1900 
& 240K 
& 16 
& $\sim$36.2 km$^2$ \\

OpenScene~\citep{openscene2023} 
& nuPlan 
& C+L 
& \textcolor{ForestGreen}{\usym{2713}} 
& 1.8K 
& 0.4M 
& 2 
& $\sim$400--500 km$^2$ \\

Nuplan-Occ~\citep{li2025scaling} 
& nuPlan 
& C+L 
& \textcolor{ForestGreen}{\usym{2713}} 
& \textbf{19K} 
& \textbf{3.6M} 
& 12 
& \textbf{$\sim$400--500 km$^2$} \\

\midrule
\rowcolor[HTML]{EFEFEF} 
\multicolumn{8}{c}{\textbf{Indoor / Embodied AI \& Robotics}} \\
\midrule

MonoScene-NYUv2~\citep{cao2022monoscene} 
& NYUv2 
& C+D 
& \textcolor{red}{\usym{2717}} 
& 0.5K 
& 1.4K 
& 12 
& -- \\

Occ-ScanNet~\citep{yu2024iso} 
& ScanNet 
& C+Depth 
& \textcolor{red}{\usym{2717}} 
& 681 
& 65.5K 
& 12 
& $\sim$46.3K m$^2$ \\

EmbodiedOcc-ScanNet~\citep{wu2025embodiedocc} 
& Occ-ScanNet, ScanNet 
& C+Depth 
& \textcolor{red}{\usym{2717}} 
& 681 
& 85.5K 
& 12 
& $\sim$46.3K m$^2$ \\

Humanoid Occupancy~\citep{cui2025humanoidocc} 
& Self-captured 
& C+L 
& \textcolor{ForestGreen}{\usym{2713}} 
& 200 
& 40K 
& 13 
& -- \\

\midrule

\textbf{\datasetname (Ours)} 
& ScanNet, ScanNet++, Matterport3D
& \textbf{C+Depth+L} 
& \textcolor{ForestGreen}{\usym{2713}} 
& \textbf{2K+} 
& \textbf{$\sim$300K} 
& \textbf{12} 
& \textbf{\makecell{
$\sim$142.6K m$^2$ total\\
$\sim$46.3K / $\sim$31.6K / $\sim$64.7K m$^2$
}} \\

\bottomrule
\end{tabular}
}
\label{tab:datasets}
\end{table*}

\section{Related Work}
\subsection{Semantic Occupancy Representations}
Voxel-based methods predict occupied regions and semantic categories from depth or RGB~\citep{song2017sscnet,cao2022monoscene,yu2024iso}, using learned lifting and structured features~\citep{li2023voxformer,zhang2023occformer,li2023bridging,li2025sliceocc}.
In autonomous driving, occupancy models incorporate surround-view observations~\citep{huang2023tpvformer,wei2023surroundocc} and temporal reasoning~\citep{ma2024cam4docc,chen2025casualocc,li2026hierarchical}, while self-supervised and generative approaches explore additional formulations for visual scene modeling~\citep{huang2024selfocc,zhang2025occnerf,li2025scaling,li2025omninwm}.
Building on 3D Gaussian Splatting~\citep{3dgs}, Gaussian occupancy methods investigate probabilistic aggregation, geometry guidance, and progressive refinement~\citep{huang2024gaussianformer,huang2024gaussianformer2,gan2024gaussianocc,zhao2026gaussianformer3d,yan2026pgocc}, as well as language-aligned semantics~\citep{zhou2026legoocc}.
Geometry priors are also used for Gaussian occupancy prediction~\citep{qian2026splatssc,zhou2026gpocc,zhou2026gpoccpp} and occupancy prediction with flexible image inputs~\citep{cao2026occany}.
Beyond representations, other work studies native 3D supervision~\citep{boeder2026shelfocc} and evidential modeling~\citep{kalble2025evocc} for occupancy learning.\looseness=-1

\subsection{Online Occupancy Mapping and Spatial Memory}
Online methods combine reconstruction with semantic completion~\citep{completescannet} or refine global Gaussian memory~\citep{wu2025embodiedocc}, incorporating geometric and uncertainty-aware updates~\citep{wang2025embodiedocc_pp,zhang2025roboocc,guo2026sgrocc}.
Incremental fusion provides another approach: GPOcc fuses Gaussian predictions over time~\citep{zhou2026gpocc}, while FreeOcc combines SLAM with open-vocabulary Gaussian mapping~\citep{jiang2026freeocc}.
Panoramic systems further extend embodied perception to multimodal sensing~\citep{cui2025humanoidocc}.
Related visual geometry models provide geometric priors~\citep{Wang2024dust3r,leroy2024mast3r,wang2025vggt,keetha2026mapanything,li2026articulated}, while recurrent and streaming states support sequential reconstruction~\citep{wang2025cut3r,wu2025point3r,chen2026ttt3r}.
\methodname incrementally builds occupancy memory from semantic Gaussian and free-space evidence, using confidence- and visibility-aware fusion and spatial submaps for memory organization.

\subsection{Benchmarks for Embodied Semantic Occupancy}
NYUv2, ScanNet, ScanNet++, and Matterport3D~\citep{silberman2012nyuv2,scannet,scannetpp,Matterport3D2017} support indoor occupancy and embodied scene understanding benchmarks~\citep{yu2024iso,completescannet,wu2025embodiedocc,wang2024embodiedscan}.
Related benchmarks address driving~\citep{wang2023openoccupancy,tian2023occ3d,li2024sscbench,openscene2023}, human-aware~\citep{kim2026mobileocc}, panoramic~\citep{cui2025humanoidocc,shi2026oneocc}, and open-vocabulary occupancy~\citep{jiang2026freeocc}.
\datasetname unifies local prediction, room-level online mapping, and building-level mapping within a hierarchical benchmark spanning perspective and panoramic observations.
Further discussion is provided in the supplementary material.

\section{GEM-Occ}

\begin{figure}[t]
    \centering
    \includegraphics[width=\linewidth]{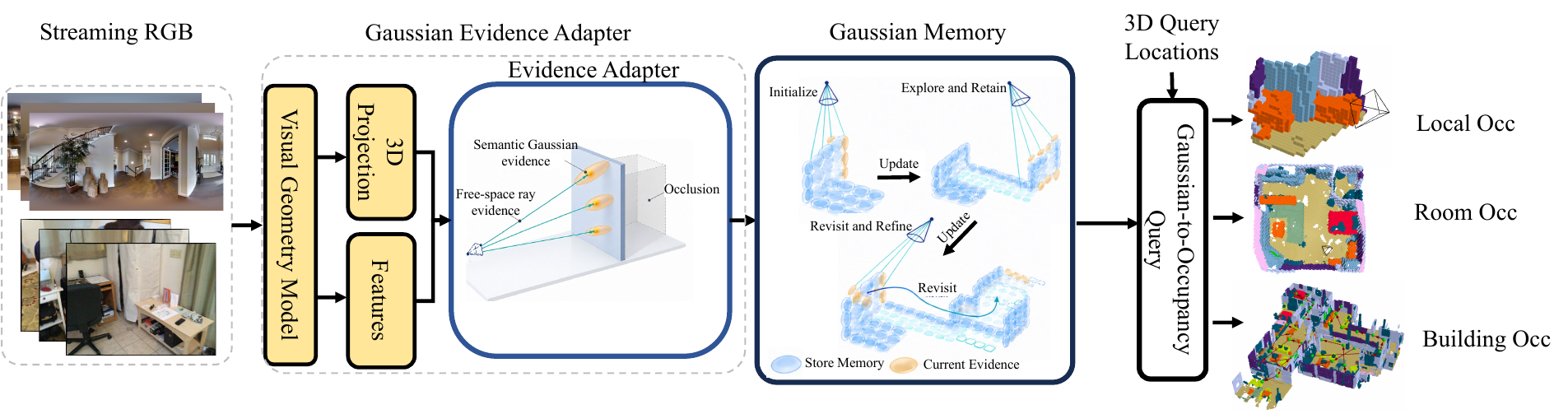}
    \caption{
    \textbf{Overview of \methodname: long-term evidence forms reliable memories.}
    Streaming perspective and panoramic observations provide geometric predictions and visual features, which the Gaussian evidence adapter converts into world-space semantic Gaussian evidence and free-space rays.
    Causal memory updates allocate newly observed structures and consolidate evidence through exploration and revisits.
    Confidence-weighted fusion reinforces supported structures, while free-space evidence revises contradicted occupancy and visibility-aware updates retain occluded memory.
    Gaussian-to-occupancy queries read out local, room-level, and building-level semantic occupancy from the accumulated memory.
    }
    \label{fig:pipeline}
\end{figure}

\methodname builds global semantic occupancy memory by allocating map elements as observations arrive and consolidating occupied and free-space evidence over time.
Fig.~\ref{fig:pipeline} shows how observation evidence, causal fusion, hierarchical maintenance, and occupancy queries support memory construction across exploration and revisits.

\subsection{Observation-Conditioned Occupancy Evidence}

Each observation contributes semantic Gaussian evidence and free-space rays for memory updates.

We use a VGGT-based encoder~\citep{wang2025vggt} to extract features $F_t$ from each posed observation, represented by a perspective frame or the calibrated sub-views of a panorama.
Prediction heads map these features to geometry $D_t$, semantic logits $S_t$, and confidence $Q_t$.
Each valid surface prediction yields
\begin{equation}
    g_i^t =
    (\mu_i^t,\Sigma_i^t,\alpha_i^t,p_i^t,\eta_i^t),
\end{equation}
where $\mu_i^t$ is the center transformed into the shared world frame using the supplied calibration and poses, and $\Sigma_i^t$ is its covariance.
The Gaussian evidence adapter predicts opacity $\alpha_i^t$ from $F_t(u)$, while $p_i^t=\mathrm{softmax}(S_t(u))$ and $\eta_i^t=Q_t(u)$ specify semantics and confidence at image location $u$.
Opacity weights local occupancy splatting; confidence weights memory fusion.

Viewing geometry and depth uncertainty determine the spatial support:
\begin{equation}
    \Sigma_i^t =
    R_i^t
    \mathrm{diag}
    \left(
        \sigma_{\perp,i}^{2},
        \sigma_{\perp,i}^{2},
        \sigma_{\parallel,i}^{2}
    \right)
    (R_i^t)^\top.
\end{equation}
Here, $R_i^t$ aligns the longitudinal axis with the world-frame viewing ray; $\sigma_{\perp,i}$ follows the projected pixel footprint, and $\sigma_{\parallel,i}$ reflects depth uncertainty.
Free-space evidence extends along each valid ray before its predicted surface hit, allowing visible contradictions to revise occupancy while preserving occluded memory.
Further details on observation processing, covariance scale construction, and parameter settings are provided in the supplementary material.

\subsection{Incremental Memory Allocation and Causal Fusion}

The memory $\mathcal{M}_t=(\mathcal{G}^{\mathrm{occ}}_t,\mathcal{R}^{\mathrm{free}}_t)$ stores occupied Gaussians and a sparse free-space evidence cache.
Each memory Gaussian contains
\begin{equation}
    G_j^t =
    (\mu_j^t,\Sigma_j^t,\ell_j^t,p_j^t,w_j^t,n_j^t),
\end{equation}
where $\ell_j^t$ is occupancy log-odds, $w_j^t$ is accumulated evidence weight, and $n_j^t$ counts supporting observations.
Incoming evidence is fused with the closest nearby primitive whose squared Mahalanobis distance, computed using the memory covariance, is below $\tau_m$; unmatched evidence allocates a new primitive.
For a matched pair, confidence-weighted fusion updates geometry and semantics:
\begin{equation}
\begin{aligned}
    \bar{w}_j^t
    &= \lambda w_j^{t-1}+\eta_i^t, \\
    \mu_j^t
    &= \frac{
        \lambda w_j^{t-1}\mu_j^{t-1}
        +\eta_i^t\mu_i^t
    }{\bar{w}_j^t}, \\
    p_j^t
    &= \frac{
        \lambda w_j^{t-1}p_j^{t-1}
        +\eta_i^t p_i^t
    }{\bar{w}_j^t}.
\end{aligned}
\end{equation}
Here, $\lambda\in[0,1]$ discounts historical evidence and $w_j^t=\bar{w}_j^t$.
Covariance follows a weighted moment merge that accounts for both primitive covariances and their center displacement.

Supporting surfaces and contradictory free-space observations update occupancy:
\begin{equation}
    \ell_j^t =
    \ell_j^{t-1}
    +\eta_i^t\Delta\ell_{\mathrm{occ}}
    -\beta_j^t\Delta\ell_{\mathrm{free}},
\end{equation}
where $\Delta\ell_{\mathrm{occ}},\Delta\ell_{\mathrm{free}}>0$ are the log-odds increments, and $\beta_j^t\in[0,1]$ measures overlap of the primitive's spatial support with currently observed free-space segments.
The occupied term is zero without a surface match; only visible segments before predicted hits contribute to $\beta_j^t$ and are accumulated in $\mathcal{R}^{\mathrm{free}}_t$.
These causal updates reinforce supported structures, suppress contradicted occupancy, and retain occluded memory.
Association and fusion details, including parameter settings and free-space overlap computation, are provided in the supplementary material.

\subsection{Hierarchical Memory Organization and Maintenance}

We organize memory into a local cache, persistent spatial submaps, and a building-level connectivity graph.
World-space indexing allocates submaps as new regions are observed and assigns observations by camera pose; revisits update the stored Gaussians through causal fusion.
Graph nodes represent submaps, with edges following the provided observation-graph connectivity.

Maintenance merges nearby primitives with overlapping covariance support and compatible semantic distributions:
\begin{equation}
    D_{\mathrm{SKL}}(p_j^t,p_k^t)
    =
    D_{\mathrm{KL}}(p_j^t\|p_k^t)
    +
    D_{\mathrm{KL}}(p_k^t\|p_j^t)
    < \tau_{\mathrm{sem}},
\end{equation}
where $\tau_{\mathrm{sem}}$ is the semantic compatibility threshold.
Merged parameters follow the confidence-weighted fusion rule in Section~3.2.
Pruning removes primitives with persistently low occupancy, high semantic entropy, or repeated free-space contradictions.
Details of submap organization and memory maintenance are provided in the supplementary material.

\subsection{Occupancy Queries and Learning Objectives}

For a query location $x$, let $\mathcal{N}_t(x)$ contain nearby memory Gaussians and $\pi_j^t=\operatorname{sigmoid}(\ell_j^t)$ denote their occupancy probabilities.
Gaussian responses yield
\begin{equation}
\begin{aligned}
    \kappa_j^t(x)
    &= \exp\!\left[
        -\frac{1}{2}(x-\mu_j^t)^\top
        (\Sigma_j^t)^{-1}(x-\mu_j^t)
    \right],\\
    P_{\mathrm{occ}}^t(x)
    &= 1-\prod_{G_j^t\in\mathcal{N}_t(x)}
    \left(1-\pi_j^t\kappa_j^t(x)\right).
\end{aligned}
\end{equation}
Semantic predictions use the same spatial and occupancy weights:
\begin{equation}
    P_{\mathrm{sem}}^t(x)
    =
    \frac{
        \sum_{G_j^t\in\mathcal{N}_t(x)}
        \pi_j^t\kappa_j^t(x)p_j^t
    }{
        \sum_{G_j^t\in\mathcal{N}_t(x)}
        \pi_j^t\kappa_j^t(x)+\epsilon
    },
\end{equation}
where $\epsilon>0$ ensures numerical stability.
A query is occupied when $P_{\mathrm{occ}}^t(x)>\tau_{\mathrm{occ}}$, with threshold $\tau_{\mathrm{occ}}\in(0,1)$ and class $\arg\max_c P_{\mathrm{sem},c}^t(x)$.
Remaining queries are free when supported by $\mathcal{R}^{\mathrm{free}}_t$, and unknown otherwise.

Training unrolls memory fusion over multiple observations and supervises memory queries together with auxiliary local predictions obtained by splatting Gaussians with their opacities $\alpha_i^t$.
The visual prediction network and evidence adapter minimize
\begin{equation}
    \mathcal{L}
    =
    \lambda_{\mathrm{occ}}\mathcal{L}_{\mathrm{occ}}
    +\lambda_{\mathrm{sem}}\mathcal{L}_{\mathrm{sem}}
    +\lambda_{\mathrm{geo}}\mathcal{L}_{\mathrm{geo}}
    +\lambda_{\mathrm{ray}}\mathcal{L}_{\mathrm{ray}},
\end{equation}
where the $\lambda$ coefficients balance the objectives.
Occupancy and semantic losses supervise valid occupied/free states and occupied-voxel semantics in both local and memory predictions.
The ray loss penalizes occupancy along observed free-space segments in both outputs, while the geometry loss supervises predicted surfaces.
Unknown targets are excluded from occupancy supervision.
Further details on occupancy readout and supervision of local and fused-memory predictions are provided in the supplementary material.

\vspace{-10pt}
\section{HIOcc and Evaluation Protocol}
\label{sec:dataset}
\vspace{-5pt}
We construct \datasetname from calibrated observations and annotated geometry in ScanNet~\citep{scannet}, ScanNet++~\citep{scannetpp}, and Matterport3D~\citep{Matterport3D2017}.
Each sample is one posed perspective frame or one panorama with its calibrated sub-views.
The shared label space in Table~\ref{tab:datasets} comprises the 11 occupied classes of Occ-ScanNet~\citep{yu2024iso} and one free-space label; unknown locations are excluded from evaluation.

\begin{figure}[h]
    \centering
    \vspace{-10pt}
    \includegraphics[width=0.85\linewidth]{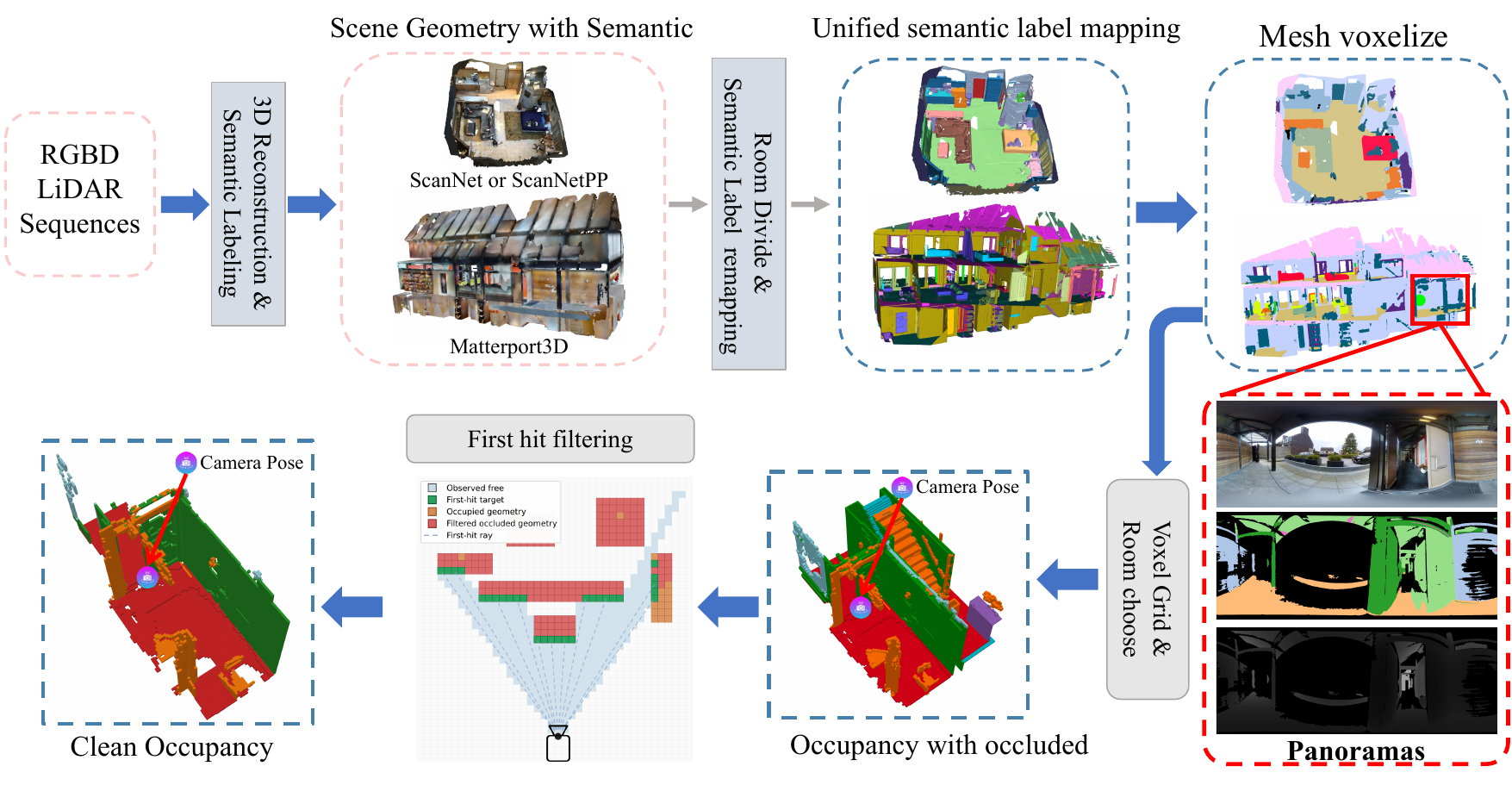}
    \vspace{-10pt}
    \caption{
    \textbf{\datasetname annotation pipeline.}
    Annotated scene geometry is mapped to shared semantic labels and converted into sparse occupancy targets through viewpoint-dependent cropping and filtering.
    }
    \label{fig:indoor_occ_dataset_pipeline}
    \vspace{-10pt}
\end{figure}

As shown in Fig.~\ref{fig:indoor_occ_dataset_pipeline}, annotation construction maps source categories to the shared taxonomy, voxelizes scene geometry, and crops and filters targets for each viewpoint.
Perspective targets use frustum and depth-consistency filtering, while panoramic targets use first-hit filtering from the shared panorama center.
Table~\ref{tab:app_dataset_annotation_ablation} evaluates 2D--3D semantic consistency between projected occupied voxels and image-level labels.
The full pipeline, including multiview semantic validation, raises semantic-consistency mIoU from $56.5$ for mesh voxelization alone to $74.9$.
Target formats, grid specifications, and filtering details are provided in the supplementary material.

\vspace{-5pt}
\begin{table*}[h]
\centering
\caption{
\textbf{Ablation of \datasetname annotation construction.}
Entries report 2D--3D semantic-consistency IoU for the 11 occupied classes.
\textbf{Complete} denotes a complete-geometry voxel reference, e.g., CompleteScanNet~\citep{completescannet};
\textbf{Mesh} denotes the mesh-based semantic voxel source.
\textbf{Ray-depth} denotes depth-consistency checks for perspective
observations and first-hit filtering for panoramic observations.
\textbf{MV Sem.} denotes multiview semantic validation.
}
\vspace{-5pt}
\label{tab:app_dataset_annotation_ablation}
\small
\setlength{\tabcolsep}{2.5pt}
\resizebox{\textwidth}{!}{
\begin{tabular}{cccccc|ccccccccccc|c}
\toprule
\textbf{Complete} & \textbf{Mesh} & \textbf{Crop} & \textbf{Frustum} & \textbf{Ray-depth} & \textbf{MV Sem.}
& \textbf{ceiling} & \textbf{floor} & \textbf{wall} & \textbf{window} & \textbf{chair} & \textbf{bed} & \textbf{sofa} & \textbf{table} & \textbf{TV} & \textbf{furniture} & \textbf{objects} & \textbf{mIoU} \\
\midrule
\checkmark & -- & -- & -- & -- & --
& 79.0 & 88.0 & 78.5 & 55.5 & 68.0 & 77.0 & 72.5 & 63.0 & 51.5 & 69.0 & 63.5
& 69.6 \\
-- & \checkmark & -- & -- & -- & --
& 72.4 & 82.1 & 66.5 & 39.7 & 53.9 & 62.4 & 57.8 & 47.8 & 35.6 & 52.7 & 50.6
& 56.5 \\
-- & \checkmark & \checkmark & -- & -- & --
& 73.8 & 83.6 & 68.1 & 41.2 & 55.2 & 64.0 & 59.1 & 49.4 & 37.1 & 54.0 & 52.3
& 58.0 \\
-- & \checkmark & \checkmark & \checkmark & -- & --
& 80.6 & 87.9 & 74.6 & 49.8 & 62.8 & 70.7 & 66.2 & 57.1 & 44.3 & 61.5 & 58.9
& 64.9 \\
-- & \checkmark & \checkmark & \checkmark & \checkmark & --
& 86.9 & 91.2 & 81.7 & 58.6 & 70.5 & 78.9 & 74.4 & 65.8 & 53.2 & 69.8 & 67.4
& 72.6 \\
-- & \checkmark & \checkmark & \checkmark & \checkmark & \checkmark
& 88.2 & 92.5 & 83.4 & 61.3 & 73.1 & 81.0 & 76.8 & 68.6 & 56.4 & 72.1 & 70.2
& \textbf{74.9} \\
\bottomrule
\end{tabular}
}
\end{table*}

Local prediction and room-level mapping use ScanNet and ScanNet++, with single posed RGB observations and causal perspective sequences, respectively.
Building-level mapping uses connected panoramic viewpoints across rooms in Matterport3D.
Methods within each track share target grids, semantic labels, valid masks, and observation protocols.
We report occupancy IoU and semantic mIoU over the 11 occupied classes, with online mapping additionally evaluated through progress AUC, revisit consistency, memory footprint, and query latency.
Progress AUC measures the area under the occupancy-accuracy curve over exploration progress.
Revisit consistency measures prediction agreement in previously observed regions upon revisiting.
HIOcc data are available on
\href{https://huggingface.co/datasets/zhuhu00/HI-Occ}{Hugging Face}.

\section{Experiments}
\label{sec:experiments}

We evaluate the quality of local occupancy evidence, its integration into persistent room-level maps, and the accuracy--efficiency trade-off during building-level exploration.
Component ablations examine how evidence representation, free-space constraints, confidence weighting, and memory maintenance contribute to these results.

\paragraph{Experimental Setup.}
We follow the three evaluation tracks defined in Section~\ref{sec:dataset}.
Local and room-level experiments use the ScanNet and ScanNet++ perspective-image splits, while building-level experiments use panoramic sequences from Matterport3D.
For the building-level setting, we train the evidence encoder and Gaussian evidence adapter on Matterport3D with supervision on fused-memory outputs and auxiliary supervision on local predictions.
Methods within each track share semantic labels, target grids, valid masks, and observation protocols.
We report occupancy IoU and semantic mIoU, together with progress AUC, revisit consistency, memory footprint, and query latency for online mapping.
Local baselines include MonoScene~\citep{cao2022monoscene}, ISO~\citep{yu2024iso}, SplatSSC~\citep{qian2026splatssc}, EmbodiedOcc~\citep{wu2025embodiedocc}, EmbodiedOcc++~\citep{wang2025embodiedocc_pp}, and GPOcc~\citep{zhou2026gpocc}.
Room-level comparisons additionally include SplicingOcc~\citep{wu2025embodiedocc}.
Further experimental settings and robustness evaluations are provided in the supplementary material.\looseness=-1

\begin{figure}[!t]
    \centering
    \includegraphics[width=0.99\linewidth]{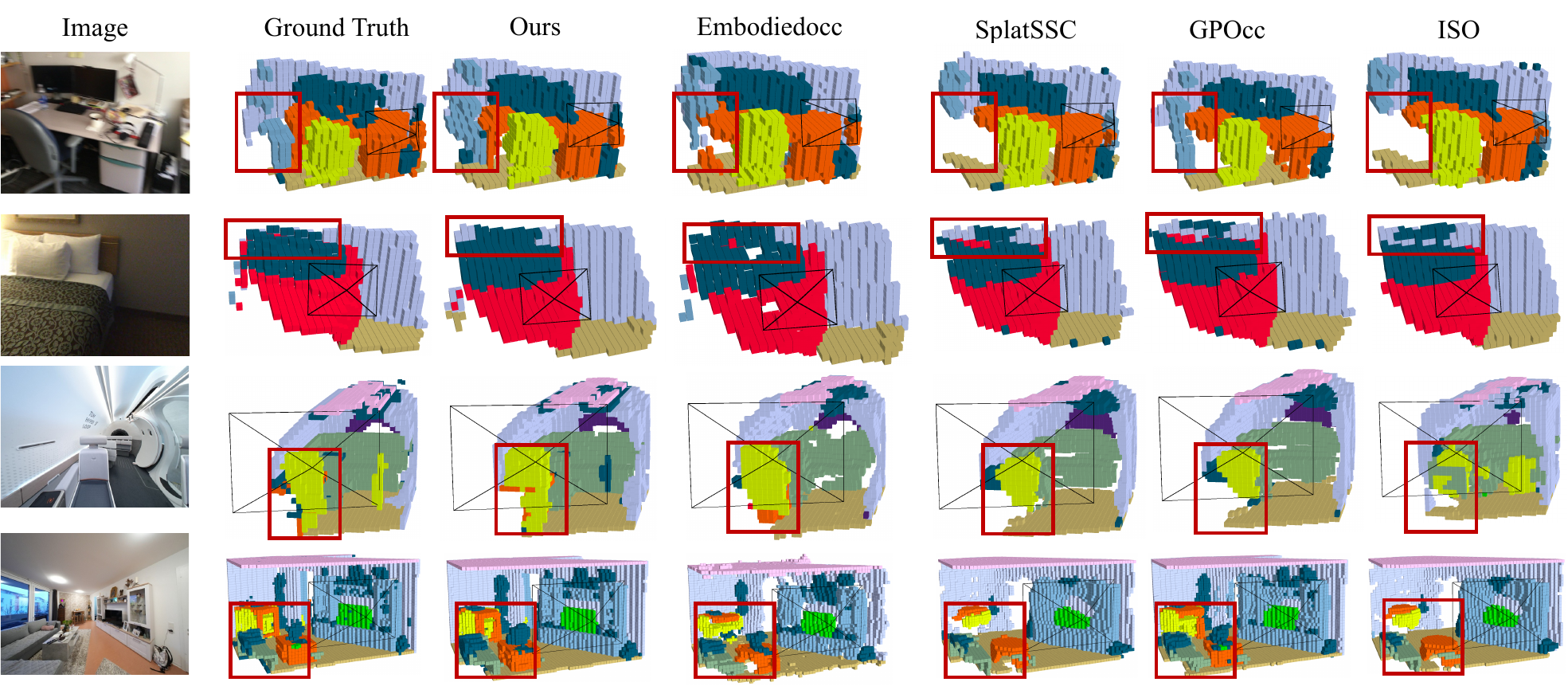}
    \caption{
    \textbf{Qualitative local occupancy prediction.}
    \methodname produces cleaner semantic structure and fewer spurious occupied regions on \datasetname compared to other methods.
    } 
    \vspace{-10pt}
    \label{fig:occ_comparison}
\end{figure}

\paragraph{Local Semantic Occupancy Prediction.}
\methodname improves both occupancy recovery and semantic prediction from individual observations.
As shown in Table~\ref{tab:app_local_prediction}, it achieves $61.37$ IoU and $57.76$ mIoU, exceeding GPOcc by $0.68$ and $2.68$, respectively.
The improvement covers all 11 semantic categories, with the largest gain on ceilings ($+20.49$) and additional gains on chairs ($+2.40$) and sofas ($+1.59$).
These results indicate that the benefit extends across structural surfaces and object categories, providing more accurate semantic evidence for subsequent map construction.
Figure~\ref{fig:occ_comparison} illustrates this behavior: the highlighted regions show more complete structures where several baseline predictions contain missing or fragmented occupancy.
The contribution of the evidence representation is further examined through the pointmap and semantic-Gaussian ablations below.

\begin{table*}[t]
\centering
\caption{
\textbf{Local semantic occupancy prediction on \datasetname.}
We report occupancy IoU and per-class semantic IoU from a single posed RGB observation.
}
\label{tab:app_local_prediction}
\scriptsize
\setlength{\tabcolsep}{2.6pt}
\resizebox{\textwidth}{!}{
\begin{tabular}{l|c|ccccccccccc|c}
\toprule
\textbf{Method}
& \textbf{IoU}
& \rotatebox{90}{\textcolor{occ_ceiling}{$\blacksquare$} ceiling}
& \rotatebox{90}{\textcolor{occ_floor}{$\blacksquare$} floor}
& \rotatebox{90}{\textcolor{occ_wall}{$\blacksquare$} wall}
& \rotatebox{90}{\textcolor{occ_window}{$\blacksquare$} window}
& \rotatebox{90}{\textcolor{occ_chair}{$\blacksquare$} chair}
& \rotatebox{90}{\textcolor{occ_bed}{$\blacksquare$} bed}
& \rotatebox{90}{\textcolor{occ_sofa}{$\blacksquare$} sofa}
& \rotatebox{90}{\textcolor{occ_table}{$\blacksquare$} table}
& \rotatebox{90}{\textcolor{occ_tvs}{$\blacksquare$} TV}
& \rotatebox{90}{\textcolor{occ_furniture}{$\blacksquare$} furniture}
& \rotatebox{90}{\textcolor{occ_objects}{$\blacksquare$} objects}
& \textbf{mIoU} \\
\midrule
MonoScene~\citep{cao2022monoscene}
& 50.54 & 74.81 & 62.33 & 35.28 & 35.74 & 32.07 & 44.45 & 39.27 & 42.66 & 34.00 & 32.06 & 26.64 & 41.76 \\
ISO~\citep{yu2024iso}
& 52.75 & 74.91 & 63.89 & 37.06 & 33.67 & 32.76 & 43.09 & 37.65 & 44.74 & 34.73 & 33.42 & 27.49 & 42.13 \\
SplatSSC~\citep{qian2026splatssc}
& 58.15 & 67.69 & 68.09 & 46.21 & 44.61 & 38.24 & 52.50 & 46.08 & 47.53 & 42.69 & 39.14 & 34.53 & 47.94 \\
EmbodiedOcc~\citep{wu2025embodiedocc}
& 53.58 & 64.33 & 70.73 & 38.21 & 38.26 & 34.23 & 46.22 & 42.91 & 44.87 & 34.95 & 36.12 & 30.13 & 43.72 \\
EmbodiedOcc++~\citep{wang2025embodiedocc_pp}
& 55.62 & 64.91 & 71.23 & 39.18 & 39.32 & 34.59 & 46.81 & 43.29 & 45.14 & 35.62 & 36.63 & 30.34 & 44.28 \\
GPOcc~\citep{zhou2026gpocc}
& 60.69 & 65.96 & 76.83 & 52.38 & 52.08 & 45.29 & 58.95 & 55.43 & 56.51 & 53.62 & 47.97 & 40.85 & 55.08 \\
\methodname
& \textbf{61.37} & \textbf{86.45} & \textbf{77.31} & \textbf{52.92} & \textbf{52.46} & \textbf{47.69} & \textbf{59.77} & \textbf{57.02} & \textbf{57.40} & \textbf{54.64} & \textbf{48.37} & \textbf{41.37} & \textbf{57.76} \\
\bottomrule
\end{tabular}
}
\end{table*}
\paragraph{Room-level Online Occupancy Mapping.}
The accuracy advantage is retained when observations are integrated into persistent room-scale maps.
Table~\ref{tab:app_room_mapping} reports $56.79$ IoU and $46.20$ mIoU for \methodname, improving over GPOcc by $3.85$ and $1.39$, respectively.
Compared with EmbodiedOcc, semantic mIoU increases by $5.89$.
Improvements over GPOcc occur across all 11 categories, including walls ($+4.07$) and chairs ($+2.85$), showing that both structural and object-level predictions benefit in the accumulated map.
In \methodname, supporting observations refine existing estimates, while observed free space can revise contradicted occupancy and confidence weighting controls the influence of incoming evidence.
The free-space and confidence ablations below support the contribution of these updates to room-level accuracy and revisit consistency.

\begin{table*}[t]
\centering
\caption{
\textbf{Room-level online occupancy mapping on \datasetname.}
All methods follow the same causal observation protocol. 
We report occupancy IoU and per-class semantic IoU after online fusion.
}
\label{tab:app_room_mapping}
\scriptsize
\setlength{\tabcolsep}{2.6pt}
\resizebox{\textwidth}{!}{
\begin{tabular}{l|c|ccccccccccc|c}
\toprule
\textbf{Method}
& \textbf{IoU}
& \rotatebox{90}{\textcolor{occ_ceiling}{$\blacksquare$} ceiling}
& \rotatebox{90}{\textcolor{occ_floor}{$\blacksquare$} floor}
& \rotatebox{90}{\textcolor{occ_wall}{$\blacksquare$} wall}
& \rotatebox{90}{\textcolor{occ_window}{$\blacksquare$} window}
& \rotatebox{90}{\textcolor{occ_chair}{$\blacksquare$} chair}
& \rotatebox{90}{\textcolor{occ_bed}{$\blacksquare$} bed}
& \rotatebox{90}{\textcolor{occ_sofa}{$\blacksquare$} sofa}
& \rotatebox{90}{\textcolor{occ_table}{$\blacksquare$} table}
& \rotatebox{90}{\textcolor{occ_tvs}{$\blacksquare$} TV}
& \rotatebox{90}{\textcolor{occ_furniture}{$\blacksquare$} furniture}
& \rotatebox{90}{\textcolor{occ_objects}{$\blacksquare$} objects}
& \textbf{mIoU} \\
\midrule
SplicingOcc~\citep{wu2025embodiedocc}
& 44.78 & 53.12 & 62.34 & 36.41 & 33.72 & 32.12 & 43.98 & 36.64 & 43.71 & 34.12 & 33.28 & 27.43 & 39.72 \\
EmbodiedOcc~\citep{wu2025embodiedocc}
& 45.12 & 52.21 & 62.41 & 36.74 & 34.81 & 32.09 & 44.82 & 37.41 & 44.81 & 35.86 & 33.73 & 28.52 & 40.31 \\
EmbodiedOcc++~\citep{wang2025embodiedocc_pp}
& 46.45 & 53.41 & 64.43 & 37.91 & 33.82 & 32.89 & 44.93 & 38.19 & 42.03 & 35.79 & 33.75 & 28.82 & 40.54 \\
GPOcc~\citep{zhou2026gpocc}
& 52.94 & 53.82 & 66.71 & 43.96 & 45.01 & 36.93 & 48.98 & 42.33 & 45.98 & 38.91 & 36.34 & 33.93 & 44.81 \\
\methodname
& \textbf{56.79} & \textbf{54.93} & \textbf{66.89} & \textbf{48.03} & \textbf{46.89} & \textbf{39.78} & \textbf{49.21} & \textbf{43.04} & \textbf{47.13} & \textbf{39.17} & \textbf{37.31} & \textbf{35.79} & \textbf{46.20} \\
\bottomrule
\end{tabular}
}
\end{table*}

\begin{figure}[h]
    \centering
    \includegraphics[width=0.99\linewidth]{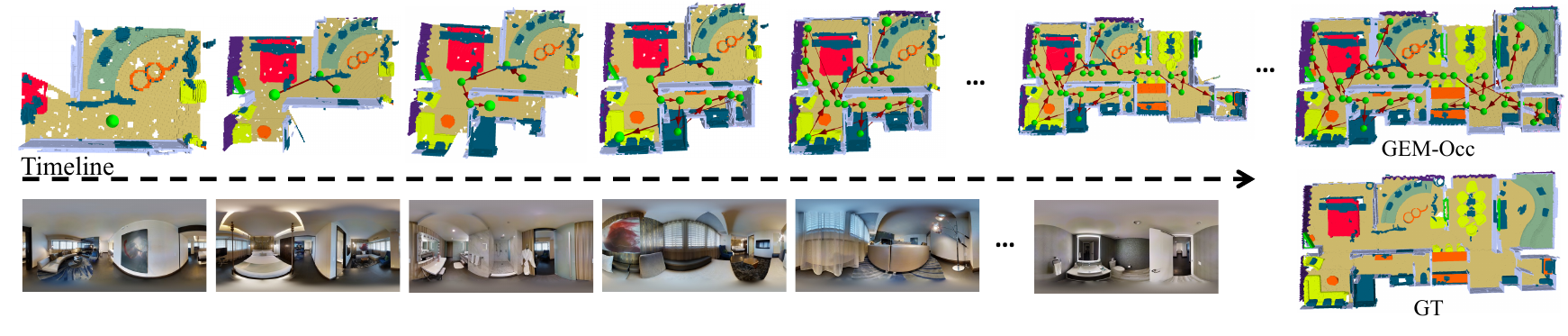}
    \vspace{-10pt}
    \caption{
    \textbf{Streaming building-level (multi-room) mapping.}
    \methodname incrementally builds semantic occupancy over connected panoramic observations.
    } 
    \label{fig:streaming_building}
\end{figure}

\paragraph{Building-level Occupancy Mapping.}
Building-level evaluation examines whether accurate and consistent memory can be maintained as exploration extends across connected rooms.
The variants in Table~\ref{tab:building_main} share the same Matterport-trained predictor and supervision, allowing comparison of different memory representations and update strategies.
Relative to flat Gaussian memory, \methodname raises mIoU from $43.1$ to $46.7$, progress AUC from $37.2$ to $41.5$, and revisit consistency from $80.8$ to $86.2$.
The higher progress AUC indicates better accumulated map quality over the course of exploration, while the consistency improvement shows more stable predictions when previously observed regions are revisited.
Figure~\ref{fig:streaming_building} illustrates how the map expands as additional panoramic viewpoints expose connected regions.\looseness=-1

These quality improvements accompany lower storage and access costs than flat Gaussian memory.
Memory per explored square meter decreases from $1.36$ to $0.81$ MB/m$^2$, and dense-query latency decreases from $49.5$ to $33.8$ ms, corresponding to reductions of $40.4\%$ and $31.7\%$.
Thus, the organized memory provides higher mapping quality with a smaller footprint and lower query cost in this comparison.
Post-memory fusion uses slightly less memory and query time, but achieves lower mIoU and revisit consistency.
The full model therefore offers a different accuracy--efficiency trade-off, with its gains in persistent-map quality supported by the memory-organization and maintenance ablations below.\looseness=-1

\begin{table*}[t]
\centering
\caption{
\textbf{Building-level occupancy mapping on Matterport3D.}
Memory is reported per explored square meter; query latency is measured for dense occupancy queries.
}
\label{tab:building_main}
\small
\resizebox{1.0\linewidth}{!}{
\begin{tabular}{lcccccc}
\toprule
Method 
& IoU
& mIoU
& Progress AUC
& Revisit Cons.
& Mem. MB/m$^2$
& Query Lat. / ms \\
\midrule
Per-frame fusion 
& 38.6 & 31.8 & 24.7 & 62.4 & 0.42 & 18.3 \\
Pointmap fusion 
& 42.9 & 35.6 & 28.9 & 68.7 & 0.58 & 24.6 \\
Post-memory fusion 
& 48.7 & 41.3 & 35.8 & 76.5 & 0.73 & 31.2 \\
Flat Gaussian memory 
& 50.4 & 43.1 & 37.2 & 80.8 & 1.36 & 49.5 \\
\methodname 
& 53.8 & 46.7 & 41.5 & 86.2 & 0.81 & 33.8 \\
\bottomrule
\end{tabular}
}
\end{table*}

\paragraph{Ablation Studies.}
Table~\ref{tab:method_ablation} first examines the representation used to construct observation evidence.
Replacing Gaussian evidence with pointmap voxelization reduces local mIoU from $57.76$ to $53.31$ and room-level mIoU from $46.20$ to $44.72$, while revisit consistency drops from $88.6$ to $76.8$.
The simultaneous losses in local prediction and revisit consistency support the use of Gaussian evidence for both observation-level estimation and persistent mapping.
Removing semantic Gaussian evidence produces the largest local mIoU decrease among the evaluated variants, to $47.12$, and also lowers room-level mIoU to $43.08$.
This highlights the importance of the semantic evidence representation before and after temporal integration.

Free-space constraints and confidence weighting affect how observations modify the accumulated memory.
Without free-space ray evidence, room-level mIoU falls to $42.62$ and revisit consistency to $78.9$.
This behavior is consistent with the role of free-space rays in revising stored occupancy when later observations provide contradictory evidence.
Removing confidence-weighted fusion lowers room-level mIoU to $42.83$ and revisit consistency to $81.7$, while the reported memory footprint remains $0.82$ MB/m$^2$.
The quality difference at the same storage cost supports weighting incoming evidence when updating existing estimates.

Memory organization and maintenance determine how much stored information is needed to sustain mapping quality.
Removing hierarchical organization increases memory from $0.82$ to $1.43$ MB/m$^2$ while reducing room-level mIoU from $46.20$ to $43.76$.
Disabling pruning and merging more than doubles memory usage to $1.78$ MB/m$^2$, yet yields slightly lower room-level mIoU ($45.82$) and revisit consistency ($87.3$).
The additional stored primitives therefore provide no accuracy benefit in this ablation.
These results support consolidation and selective maintenance as a means of retaining a compact memory while preserving useful geometric and semantic evidence.

\begin{table}[h]
\centering
\caption{
\textbf{Ablation of \methodname.}
Local metrics are computed from single-frame predictions; room-level metrics are computed after causal fusion.
Revisit consistency and memory are measured in the room-level mapping setting, with memory reported per explored square meter.
}
\label{tab:method_ablation}
\small
\resizebox{\linewidth}{!}{
\begin{tabular}{lcc|cccc}
\toprule
Setting 
& Local IoU  
& Local mIoU  
& Room IoU  
& Room mIoU  
& Revisit Cons.  
& Mem. MB/m$^2$ \\
\midrule
Full \methodname 
& 61.37 & 57.76 
& 56.79 & 46.20 & 88.6 & 0.82 \\
w/ Pointmap voxelization 
& 57.84 & 53.31 
& 50.96 & 44.72 & 76.8 & 1.05 \\
w/o Semantic Gaussian evidence 
& 58.63 & 47.12 
& 52.31 & 43.08 & 80.4 & 0.84 \\
w/o Free-space ray evidence 
& 60.42 & 51.36 
& 52.74 & 42.62 & 78.9 & 0.80 \\
w/o Confidence-weighted fusion 
& 60.91 & 52.02 
& 54.18 & 42.83 & 81.7 & 0.82 \\
w/o Hierarchical memory 
& 61.25 & 52.41 
& 55.02 & 43.76 & 83.5 & 1.43 \\
w/o Pruning/merging 
& 61.31 & 52.48 
& 56.21 & 45.82 & 87.3 & 1.78 \\
\bottomrule
\end{tabular}
}
\end{table}

\section{Conclusion}
We presented \methodname, a Gaussian Evidence Memory framework that turns sequential visual evidence into persistent semantic occupancy memory.
The framework constructs occupied semantic Gaussians and free-space rays from each observation, allocates memory as new regions are explored, and consolidates supporting and conflicting evidence through confidence- and visibility-aware causal updates.
Hierarchical organization supports continued mapping and occupancy queries across connected indoor spaces.
We introduced \datasetname as a unified benchmark for evaluating local prediction, online room mapping, and building-level exploration under shared semantic labels and evaluation protocols.
Experiments show improved semantic occupancy accuracy and revisit consistency over the evaluated mapping baselines, together with lower memory consumption and query latency than flat Gaussian memory in building-level mapping.
These results demonstrate how evidence accumulated through exploration and revisits supports reliable and scalable spatial memory.

\input{arxiv_statements}

\bibliography{ref}
\bibliographystyle{iclr2027_conference}

\clearpage
\appendix
\input{arxiv_supplementary}

\end{document}

%% file: arxiv_preamble.tex
\usepackage{amsmath,amsfonts,bm}
\usepackage[table,dvipsnames]{xcolor}
\usepackage{hyperref}
\usepackage{url}
\usepackage[utf8]{inputenc}
\usepackage[T1]{fontenc}
\usepackage{url}
\usepackage{booktabs}
\usepackage{amsfonts}
\usepackage{nicefrac}
\usepackage{microtype}
\usepackage{amsmath}
\usepackage{amssymb}
\usepackage{mathrsfs}
\usepackage{bbm}
\usepackage{graphicx}
\usepackage{adjustbox}
\usepackage{tabularx}
\usepackage{tabularray}
\usepackage{array}
\usepackage{multirow}
\usepackage{makecell}
\usepackage{wrapfig}
\usepackage{booktabs}
\usepackage{colortbl}
\usepackage{algorithm}
\usepackage{algpseudocode}
\usepackage{xspace}
\usepackage{utfsym}
\usepackage{bbding}
\usepackage{enumitem}
\usepackage{pifont}
\usepackage{pgffor}
\usepackage{svg}
\usepackage{caption}
\usepackage{fontawesome5}

\definecolor{occ_ceiling}{RGB}{214,38,40}
\definecolor{occ_floor}{RGB}{43,160,4}
\definecolor{occ_wall}{RGB}{158,216,229}
\definecolor{occ_window}{RGB}{114,158,206}
\definecolor{occ_chair}{RGB}{204,204,91}
\definecolor{occ_bed}{RGB}{255,186,119}
\definecolor{occ_sofa}{RGB}{147,102,188}
\definecolor{occ_table}{RGB}{30,119,181}
\definecolor{occ_tvs}{RGB}{160,188,33}
\definecolor{occ_furniture}{RGB}{255,127,12}
\definecolor{occ_objects}{RGB}{196,175,214}
\definecolor{gemresourceblue}{RGB}{36,89,143}

\newcommand{\methodname}{{GEM-Occ}\xspace}
\newcommand{\datasetname}{{HIOcc}\xspace}

%% file: arxiv_abstract.tex
Embodied agents exploring indoor environments require reliable semantic occupancy memory that persists across observations and revisits. 
Building such memory is challenging because each observation provides incomplete and uncertain geometric and semantic evidence. 
We introduce GEM-Occ, a Gaussian Evidence Memory framework that consolidates evidence accumulated over time into persistent semantic occupancy memory. Local predictions are converted into occupied semantic Gaussians and free-space ray evidence. Confidence- and visibility-aware causal updates integrate supporting observations, suppress occupancy contradicted by observed free space, and preserve previously observed structures through occlusion. A hierarchical memory organization supports continued mapping and efficient queries across connected indoor spaces. 
To evaluate this capability, we introduce HIOcc, a unified benchmark for embodied semantic occupancy memory. HIOcc establishes a shared semantic label space and evaluation framework spanning local prediction, room-level online mapping, and building-level mapping, while accommodating perspective and panoramic observations. 
Experiments on HIOcc demonstrate that GEM-Occ outperforms existing methods, enabling accurate semantic occupancy prediction and consistent online mapping across spatial scales with efficient memory usage and fast occupancy queries.

%% file: arxiv_statements.tex
\subsection*{AI use statement}

We used generative AI tools to assist with implementing code,
to improve the clarity and readability of the manuscript, and to suggest
revisions to its structure and organization. 
All research ideas, methods, and conclusions were developed by the authors. The authors bear full responsibility for the content of the manuscript, including any text generated or polished by an LLM, and have ensured that it complies with ethical guidelines and does not involve plagiarism or scientific misconduct.

\subsection*{Reproducibility Statement}
We have made every effort to ensure that our results are reproducible. The main paper and supplementary material describe the architecture, evidence construction, memory fusion, and learning objectives of GEM-Occ. Section~\ref{sec:dataset} and the supplementary material detail HIOcc construction, including semantic label mapping, occupancy target generation, and visibility filtering. Section~\ref{sec:experiments} describes the experimental setup and evaluation protocols, with additional comparisons, runtime measurements, and robustness analyses provided in the supplementary material. HIOcc is constructed from ScanNet, ScanNet++, and Matterport3D. We will release the source code, trained checkpoints, and scripts for dataset construction, training, and evaluation to support independent verification and extension of this work.\looseness=-1

%% file: arxiv_supplementary.tex
\section{Additional Experiments}
\label{app:additional_experiments}

\subsection{External Baseline Comparisons}

Table~\ref{tab:supp_external_comparisons} reports adapted baselines under the \datasetname evaluation protocol across local, room-level, and building-level tasks.
In building-level mapping, \methodname achieves $53.80$ IoU and $46.70$ mIoU, exceeding GPOcc by $3.55$ and $2.89$ points, respectively.
These comparisons complement the main-text building-level variants, which use the same Matterport-trained predictor and supervision to examine different memory representations and update strategies.
LegoOcc and FreeOcc are evaluated on the same 11 occupied classes.

\begin{table}[!htbp]
\centering
\caption{
\textbf{Additional baseline comparisons on \datasetname.}
$^\dagger$ denotes adaptation to the shared evaluation protocol.
Each task reports occupancy IoU and semantic mIoU over the 11 occupied classes; -- indicates an unreported result.
}
\label{tab:supp_external_comparisons}
\small
\setlength{\tabcolsep}{5pt}
\begin{tabular}{lcccccc}
\toprule
\textbf{Method}
& \multicolumn{2}{c}{\textbf{Local}}
& \multicolumn{2}{c}{\textbf{Room}}
& \multicolumn{2}{c}{\textbf{Building}} \\
\cmidrule(lr){2-3}\cmidrule(lr){4-5}\cmidrule(lr){6-7}
& IoU & mIoU & IoU & mIoU & IoU & mIoU \\
\midrule
EmbodiedOcc$^\dagger$~\cite{wu2025embodiedocc}
& 53.58 & 43.72 & 45.12 & 40.31 & 44.92 & 40.73 \\
GPOcc$^\dagger$~\cite{zhou2026gpocc}
& 60.69 & 55.08 & 52.94 & 44.81 & 50.25 & 43.81 \\
LegoOcc$^\dagger$~\cite{zhou2026legoocc}
& 59.33 & 20.79 & -- & -- & -- & -- \\
FreeOcc$^\dagger$~\cite{jiang2026freeocc}
& -- & -- & 30.21 & 13.85 & 28.92 & 12.56 \\
\methodname
& \textbf{61.37} & \textbf{57.76} & \textbf{56.79} & \textbf{46.20} & \textbf{53.80} & \textbf{46.70} \\
\bottomrule
\end{tabular}
\end{table}

\subsection{Runtime and Memory Accounting}

Table~\ref{tab:supp_runtime} reports throughput for evidence extraction and the complete online pipeline, measured on the same hardware.
Evidence extraction runs at $8.7$ FPS before causal fusion.
The full pipeline runs at $5.2$ FPS and includes evidence extraction, causal fusion, submap and graph maintenance, and occupancy querying.

\begin{table}[!htbp]
\centering
\caption{\textbf{Runtime of \methodname.} Throughput is measured on the same hardware for the two processing scopes.}
\label{tab:supp_runtime}
\small
\begin{tabular}{lc}
\toprule
\textbf{Processing scope} & \textbf{Throughput (FPS)} \\
\midrule
Evidence extraction & 8.7 \\
Complete online pipeline & 5.2 \\
\bottomrule
\end{tabular}
\end{table}

Memory footprint is reported in MB per explored square meter.
The main-text ablation measures memory and revisit consistency in the room-level setting, while the building-level comparison measures them on panoramic sequences across connected rooms.
The full model uses $0.82$ MB/m$^2$ in the room-level ablation and $0.81$ MB/m$^2$ in the building-level comparison.
Against flat Gaussian memory in the building-level setting, memory decreases from $1.36$ to $0.81$ MB/m$^2$ and dense-query latency from $49.5$ to $33.8$ ms, reductions of $40.4\%$ and $31.7\%$, respectively.
Dense-query latency measures occupancy readout and is distinct from the time required to process a new observation through the complete pipeline.

\subsection{Robustness to Pose and Depth Perturbations}

We evaluate sensitivity to pose and geometric evidence using controlled translation, rotation, and relative depth perturbations in the building-level setting.
Table~\ref{tab:supp_robustness} reports occupancy IoU and semantic mIoU for each perturbation level.
Increasing translation from $0.1$ to $0.2$ m reduces mIoU from $40.33$ to $34.54$; increasing yaw perturbation from $5^\circ$ to $10^\circ$ reduces it from $39.25$ to $31.57$.
Relative depth noise at $\sigma=5\%$ and $10\%$ yields $42.78$ and $31.16$ mIoU, respectively, quantifying the effect of geometric uncertainty on the accumulated memory.
The degradation shows that map quality remains sensitive to alignment and surface prediction errors, even with confidence-weighted fusion.
These tests characterize input sensitivity in the posed-mapping setting; they do not evaluate trajectory estimation or SLAM drift correction.

\begin{table}[!htbp]
\centering
\caption{\textbf{Building-level robustness under pose and depth perturbations.}}
\label{tab:supp_robustness}
\small
\begin{tabular}{lcc}
\toprule
\textbf{Setting} & \textbf{IoU} & \textbf{mIoU} \\
\midrule
Default & \textbf{53.80} & \textbf{46.70} \\
Translation $0.1$ m ($x$-axis) & 50.14 & 40.33 \\
Translation $0.2$ m ($x$-axis) & 45.76 & 34.54 \\
Rotation $5^\circ$ (yaw) & 47.29 & 39.25 \\
Rotation $10^\circ$ (yaw) & 42.51 & 31.57 \\
Relative depth noise ($\sigma=5\%$) & 44.64 & 42.78 \\
Relative depth noise ($\sigma=10\%$) & 39.22 & 31.16 \\
\bottomrule
\end{tabular}
\end{table}

\paragraph{Scope and limitations.}
The current evaluation concerns static indoor scenes with supplied camera poses and a fixed semantic taxonomy.
Building-level experiments train the evidence predictor on Matterport3D; they do not establish zero-shot transfer from perspective-image training to panoramic observations.
Dynamic-scene mapping and downstream navigation or manipulation remain outside the evaluated setting.

\subsection{Additional Mapping Visualizations}

\begin{figure}[!htbp]
    \centering
    \includegraphics[width=\linewidth]{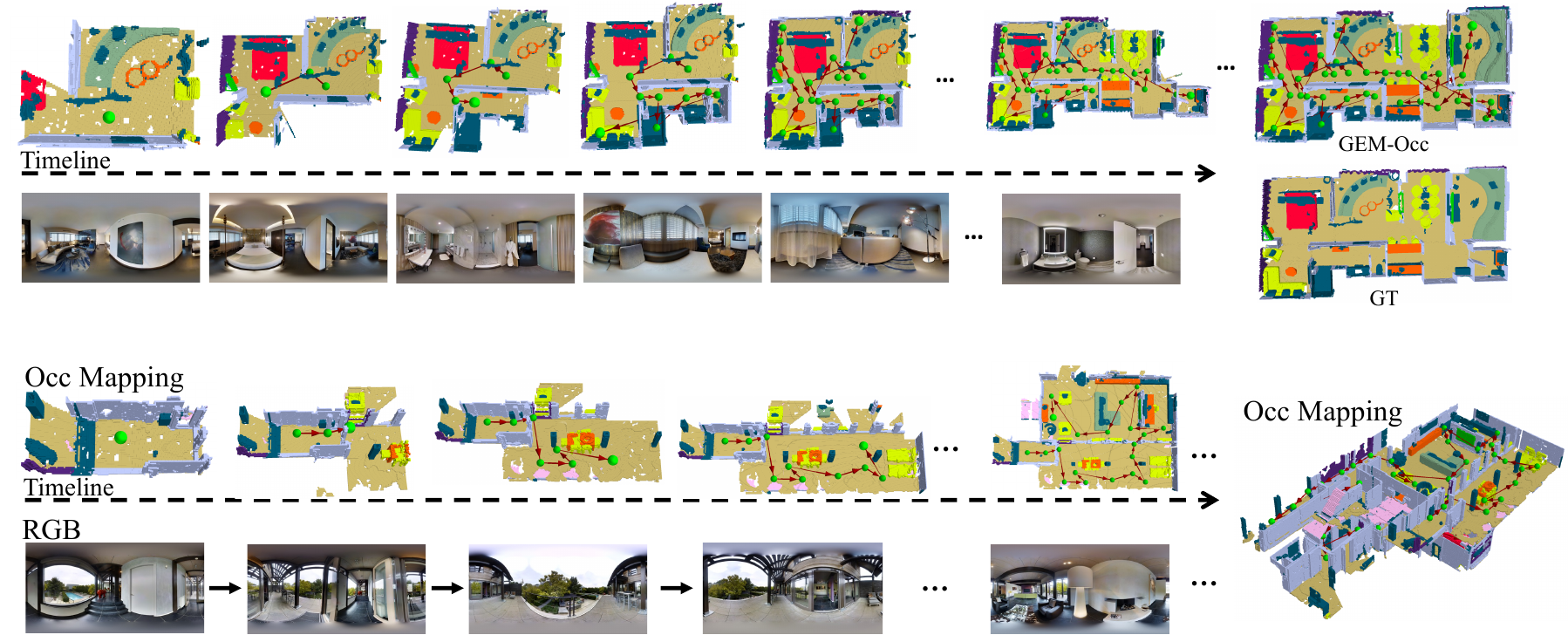}
    \caption{
    \textbf{Additional building-level mapping results.}
    Two sequences show the evolution of \methodname's semantic occupancy memory alongside the incoming panoramic RGB observations, ordered from left to right.
    The upper sequence includes the final prediction and ground-truth map on the right; the lower sequence ends with a larger view of the accumulated map.
    }
    \label{fig:supp_streaming_building}
\end{figure}

Figure~\ref{fig:supp_streaming_building} presents two additional building-level mapping sequences on Matterport3D.
As panoramic observations arrive, the accumulated map expands across connected regions while retaining previously mapped structures.
The final prediction in the upper sequence captures the major room layout and semantic regions visible in the ground-truth map.

\section{GEM-Occ Method Details}
\label{app:method_details}

\subsection{Input and Mapping Protocol}

\methodname receives posed RGB observations with camera calibration.
A time step corresponds to a perspective frame or a panorama with its calibrated sub-views, whose evidence is expressed in a shared world frame.
Depth and semantic annotations provide supervision and benchmark targets; RGB-only inference uses predicted geometry.
Camera poses are supplied, and memory updates use only observations available up to the current time step.

\subsection{Observation Evidence and Ray Support}

The adapted VGGT~\cite{wang2025vggt} encoder extracts streaming visual features from perspective and panoramic observations.
Prediction heads produce local geometry $D_t$, semantic logits $S_t$, and confidence $Q_t$; supplied calibration and poses express the predicted surfaces and viewing rays in a shared world coordinate system.
For a valid image location $u$, the Gaussian evidence adapter predicts occupancy opacity $\alpha_i^t$ from $F_t(u)$, while $p_i^t=\mathrm{softmax}(S_t(u))$ and $\eta_i^t=Q_t(u)$ provide the semantic distribution and incoming evidence confidence.
Opacity controls the contribution to local occupancy splatting, whereas confidence determines the weight of the observation during memory fusion.

The ray-aligned covariance associates each surface prediction with volumetric support determined by its viewing geometry and depth uncertainty.
Its two transverse axes follow the projected pixel footprint, and its longitudinal axis follows the world-frame viewing ray, with scale reflecting uncertainty in predicted surface depth.
Free-space evidence is recorded only on visible ray segments before the predicted surface hit.
These segments identify where the current observation can contradict stored occupancy; regions beyond the hit contribute no free-space evidence and retain their existing memory in the absence of other observations.

\subsection{Spatial Association and Covariance Fusion}

Incoming evidence is associated with memory through a local spatial search and a squared Mahalanobis-distance gate.
For an evidence primitive $g_i^t$ and a candidate $G_j^{t-1}$, the distance is
\begin{equation}
    d_M(g_i^t,G_j^{t-1})=
    (\mu_i^t-\mu_j^{t-1})^\top
    (\Sigma_j^{t-1})^{-1}
    (\mu_i^t-\mu_j^{t-1}).
\end{equation}
The closest candidate satisfying $d_M<\tau_m$ receives the incoming evidence; if none satisfies the gate, the evidence allocates a new memory primitive. An unmatched evidence primitive initializes a new memory element
with its predicted center, covariance, and semantic distribution.
Allocation therefore follows the spatial coverage of observed structures, while repeated observations consolidate existing map elements.

For a matched pair, the main-text confidence-weighted rule updates the center and semantic distribution with total weight $\bar{w}_j^t=\lambda w_j^{t-1}+\eta_i^t$.
Covariance is updated by a weighted moment merge:
\begin{equation}
\begin{aligned}
    \Sigma_j^t=\frac{1}{\bar{w}_j^t}\big[&
    \lambda w_j^{t-1}\big(\Sigma_j^{t-1}+\delta_j^{t-1}(\delta_j^{t-1})^\top\big)\\
    &+\eta_i^t\big(\Sigma_i^t+\delta_i^t(\delta_i^t)^\top\big)\big],
\end{aligned}
\end{equation}
where $\delta_j^{t-1}=\mu_j^{t-1}-\mu_j^t$ and $\delta_i^t=\mu_i^t-\mu_j^t$.
The displacement terms account for the separation of the original centers around the fused center, preserving the spatial spread of the combined evidence.

When multiple evidence primitives from the same observation are associated with one memory element, their geometric and semantic estimates are first aggregated using confidence-weighted moments, including the covariance contributions from center displacement.
The summed confidence serves as the aggregate weight.
The single-pair equations above then fuse this aggregate with the stored memory primitive.
The historical discount and free-space correction are each applied once per memory element per observation.

Occupancy updates combine the positive increment $\eta_i^t\Delta\ell_{\mathrm{occ}}$ from matched surfaces with the negative increment $\beta_j^t\Delta\ell_{\mathrm{free}}$ from observed free space.
The occupied increment is zero when there is no surface match.
Let $\mathcal{S}_j^t$ be a finite set of sample locations within the primitive's spatial support, and let $\mathcal{F}_t$
denote the discretized free-space region supported by the current observation.
We compute the overlap as
\begin{equation}
    \beta_j^t =
    \frac{
        \sum_{x\in\mathcal{S}_j^t}
        \kappa_j^t(x)\,\mathbf{1}[x\in\mathcal{F}_t]
    }{
        \sum_{x\in\mathcal{S}_j^t}\kappa_j^t(x)
    },
\end{equation}
where $\kappa_j^t(x)$ is the Gaussian spatial response defined in the main text.
This normalized overlap lies in $[0,1]$;
we set $\beta_j^t=0$ when the sample set is empty.
Only segments before predicted surface hits contribute negative evidence, so occlusion alone does not reduce stored occupancy. An existing primitive can receive a free-space correction even without a surface match in the current observation.
If neither supporting surface evidence nor contradictory free-space evidence is available, its occupancy log-odds remains unchanged.
The correction uses only the current observation's free-space evidence.
The same segments accumulate in the sparse cache $\mathcal{R}^{\mathrm{free}}_t$ for later queries.

\subsection{Submap Organization and Memory Maintenance}

World-space indexing creates spatial submaps as new regions are observed and assigns observations using their camera poses.
Each submap stores Gaussian memory in the shared world frame, allowing a revisit to update the existing primitives through the same association and fusion rules.
The room-level submaps are spatial partitions defined by the index; Matterport3D subroom annotations are used only for benchmark construction.
Graph nodes represent submaps, and edges follow the provided observation-graph connectivity.
The observation graph supplies the sequence, poses, and adjacency for this organization.
The graph records connections among mapped regions, while the spatial index supports access to their Gaussian memory for updates and queries.

Maintenance operates periodically within active submaps.
Merge candidates have nearby centers, overlapping covariance support, and compatible semantic predictions under the symmetric KL-divergence criterion in the main text.
Their parameters are combined using confidence-weighted fusion.
Pruning removes primitives whose occupancy confidence remains low after repeated observations, whose semantic entropy stays high, or that are repeatedly contradicted by free-space evidence.
Together, these operations consolidate duplicate geometry and remove poorly supported primitives as evidence accumulates.

\subsection{Occupancy Readout and Supervision}

Occupancy queries can be evaluated at arbitrary 3D locations or sampled on the voxel grid required by each evaluation setting.
Each nearby Gaussian contributes according to its spatial response and the occupancy probability obtained from its stored log-odds.
The same weights aggregate semantic predictions.
Using label $0$ for free space and $\bot$ for unknown space, the readout is
\begin{equation}
\hat y_t(x)=
\begin{cases}
\arg\max_c P_{\mathrm{sem},c}^t(x),
    & P_{\mathrm{occ}}^t(x)>\tau_{\mathrm{occ}},\\
0,  & P_{\mathrm{occ}}^t(x)\leq\tau_{\mathrm{occ}}
       \text{ and }x\in\mathcal{R}^{\mathrm{free}}_t,\\
\bot, & \text{otherwise}.
\end{cases}
\end{equation}
Occupied evidence takes precedence when Gaussian support and cached free-space support overlap.
Locations without occupied or free-space support remain unknown.

Training unrolls memory fusion over multiple observations and supervises fused-memory queries using the view-conditioned targets in \datasetname.
Local predictions are formed by splatting the incoming evidence Gaussians with their predicted opacities $\alpha_i^t$.
Auxiliary supervision on these outputs supplies a learning signal for opacity alongside the supervision on the accumulated memory.
The occupancy, semantic, and ray objectives apply to both local and fused-memory predictions.
Occupancy supervision covers valid occupied and free target locations, semantic supervision covers occupied target voxels, and ray supervision penalizes occupancy along observed free-space segments.
The geometry objective supervises predicted surfaces using the available geometric targets.
Unknown target locations are excluded from occupancy supervision.
The visual prediction network and evidence adapter are learned; the fusion and query rules introduce no additional learned parameters.
Supervision is applied to occupancy queries and predicted geometry, without Gaussian-level annotations.

\subsection{Training and Evaluation Setup}

Local and room-level experiments use the ScanNet and ScanNet++ perspective-image subsets of \datasetname.
Building-level experiments use the Matterport3D panoramic subset, with observations represented by panoramas and their calibrated sub-views.
For this setting, the evidence encoder and Gaussian evidence adapter are trained on Matterport3D using the shared memory representation and causal fusion rule.
Methods within each task are evaluated with the same semantic labels, target grids, and valid masks; online evaluation follows the same causal observation protocol.

\section{HIOcc Annotation and Target Details}
\label{app:dataset_details}

\subsection{Viewpoint Samples and Sparse Targets}

\datasetname combines calibrated observations and annotated scene geometry from ScanNet~\cite{scannet}, ScanNet++~\cite{scannetpp}, and Matterport3D~\cite{Matterport3D2017} within a shared occupancy representation.
Perspective observations cover RGB-D scans and higher-fidelity reconstructions, while panoramic observations connect rooms and floors within buildings.
A sample is one viewpoint: a posed perspective frame or a panorama with its calibrated sub-views.
Depth maps, semantic maps, auxiliary views, and targets at different resolutions support annotation and supervision for that viewpoint and are counted as part of the same sample.

Targets store sparse occupied-voxel entries $[i,j,k,l]$, where $(i,j,k)$ are voxel indices and $l$ is the semantic label, together with grid origin, voxel size, and grid shape.
The 11 occupied categories follow Occ-ScanNet~\cite{yu2024iso}: \emph{ceiling}, \emph{floor}, \emph{wall}, \emph{window}, \emph{chair}, \emph{bed}, \emph{sofa}, \emph{table}, \emph{TV}, \emph{furniture}, and \emph{objects}.
Together with the free-space label, they form the 12 labels reported for \datasetname.
The sparse files explicitly store occupied semantic labels; observation geometry and validity masks determine free and unknown states during supervision and evaluation.
Free space is supported by observed empty ray segments, while unknown locations fall outside valid target support and are masked out.
Semantic mIoU averages the 11 occupied categories, excluding free space.
An absent occupied entry does not by itself establish free space.

\subsection{Perspective and Panoramic Target Construction}

Annotation construction first maps source categories to the shared taxonomy and voxelizes annotated scene geometry in world coordinates.
For each viewpoint, a local metric volume is cropped and filtered according to its observation geometry.
For perspective targets, voxel centers are transformed into the camera frame, filtered by the image frustum, and checked for depth consistency when depth is available.
The perspective target hierarchy provides three resolutions with a common physical extent (Table~\ref{tab:supp_target_grids}).

Panoramic targets use a $120\times120\times60$ grid with $0.05$\,m voxels, covering $6\times6\times3$\,m around the shared panorama center.
For targets covered by calibrated sub-views, first-hit filtering casts rays from this center and uses ceiling, floor, wall, and window voxels as structural occluders.
The filtering retains a small region around the panorama center and a finite thickness of the first-hit structures, then removes occupancy occluded beyond that retained support.
This produces targets with volumetric structural support for panoramic observations.

\begin{table}[!htbp]
\centering
\caption{\textbf{Target grid specifications.} Perspective resolutions describe alternative targets for the same observation.}
\label{tab:supp_target_grids}
\small
\begin{tabular}{lccc}
\toprule
Target & Grid shape & Voxel size (m) & Extent (m) \\
\midrule
Perspective & $240\times240\times144$ & $0.02$ & $4.8\times4.8\times2.88$ \\
Perspective & $120\times120\times72$ & $0.04$ & $4.8\times4.8\times2.88$ \\
Perspective & $60\times60\times36$ & $0.08$ & $4.8\times4.8\times2.88$ \\
Panoramic & $120\times120\times60$ & $0.05$ & $6\times6\times3$ \\
\bottomrule
\end{tabular}
\end{table}

\begin{figure}[!htbp]
    \centering
    \includegraphics[width=\linewidth]{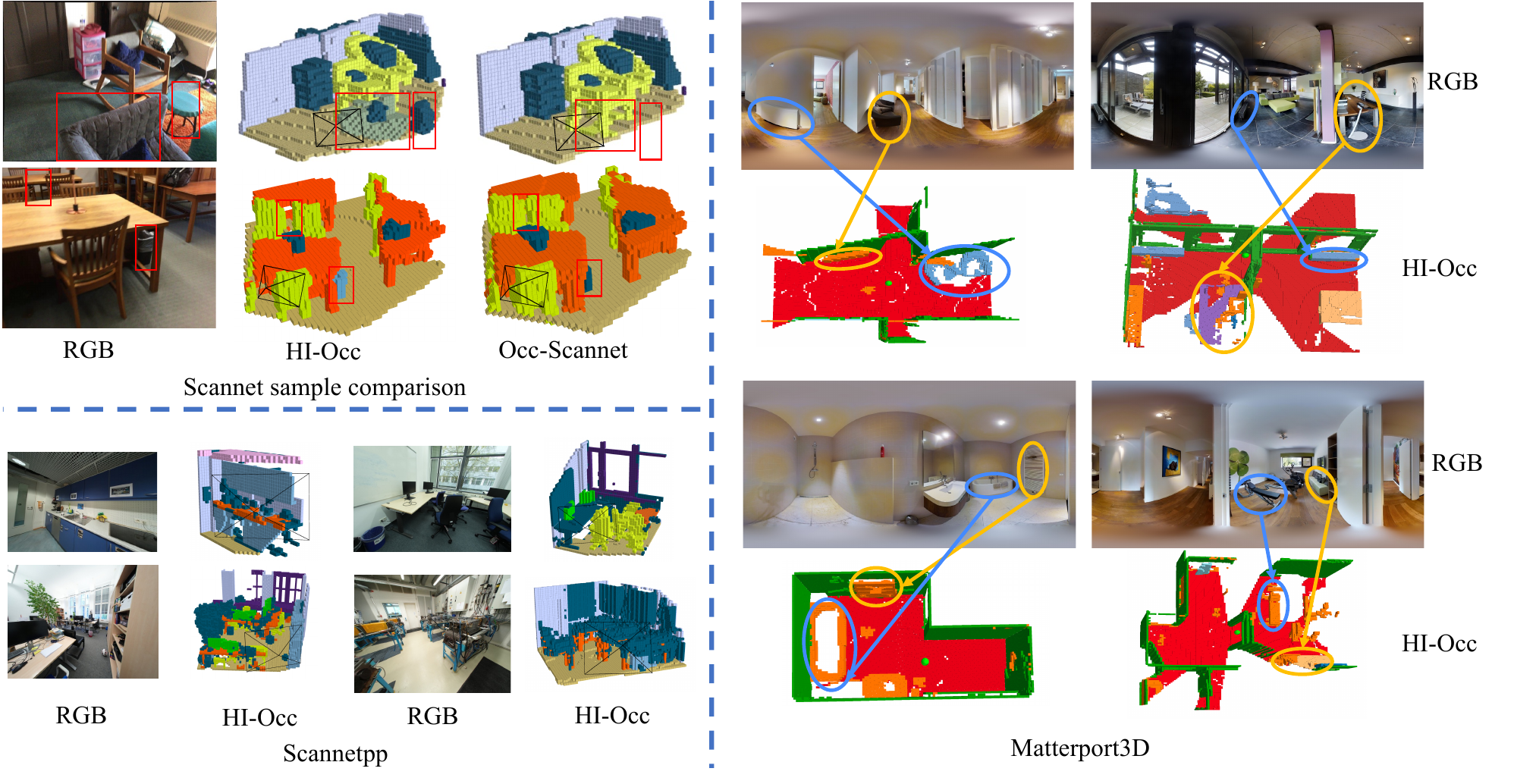}
    \caption{
    \textbf{Qualitative evaluation of \datasetname annotations.}
    Top left: comparison with Occ-ScanNet on ScanNet, with red boxes highlighting differences in annotated object structure.
    Bottom left: additional ScanNet++ examples.
    Right: Matterport3D panoramic observations and occupancy targets, with colored circles and arrows indicating corresponding structures.
    }
    \label{fig:supp_annotation_comparison}
\end{figure}

\subsection{Annotation Validation}

We validate targets through 2D--3D semantic consistency by projecting occupied voxels into calibrated views and comparing their labels with image-level semantic evidence.
The main-text ablation reports per-class consistency after mesh voxelization, local cropping, view filtering, depth or first-hit filtering, and multiview semantic validation.
The full construction raises semantic-consistency mIoU from $56.5$ to $74.9$ across the 11 occupied categories.
These scores measure agreement between constructed targets and image-level semantic evidence.

Figure~\ref{fig:supp_annotation_comparison} provides qualitative comparisons with Occ-ScanNet and additional annotation examples from ScanNet++ and Matterport3D.
In the highlighted ScanNet regions, \datasetname preserves object structures that are incomplete or missing in the Occ-ScanNet targets, improving correspondence with the RGB observations.
The remaining examples show how the annotations capture visible furniture and room structures under perspective and panoramic observation settings.

\subsection{Evaluation Protocol and Metrics}
\label{app:evaluation}

Local prediction uses a single posed RGB observation, room-level mapping uses causal perspective sequences, and building-level mapping follows connected panoramic viewpoints.
All comparisons within a track use the same target grids, labels, and valid evaluation masks.

Occupancy IoU measures overlap between predicted and target occupied voxels on valid evaluation support.
Semantic mIoU averages the class-wise intersection over union for the $C=11$ occupied categories:
\begin{equation}
    \mathrm{IoU}_{\mathrm{occ}}
    =\frac{\mathrm{TP}_{\mathrm{occ}}}
    {\mathrm{TP}_{\mathrm{occ}}+\mathrm{FP}_{\mathrm{occ}}+\mathrm{FN}_{\mathrm{occ}}},
    \qquad
    \mathrm{mIoU}=\frac{1}{C}\sum_{c=1}^{C}
    \frac{\mathrm{TP}_c}{\mathrm{TP}_c+\mathrm{FP}_c+\mathrm{FN}_c}.
\end{equation}
Free space is excluded from the semantic class average, and unknown targets are excluded by the evaluation mask.

Progress AUC summarizes the area under the occupancy-accuracy curve over exploration progress.
Revisit consistency measures prediction agreement in previously observed regions when the agent returns to them.
They complement final-map accuracy by measuring quality throughout exploration and stability upon revisiting.

\section{Extended Related Work}
\label{app:related_work}

\subsection{Semantic Occupancy Representations}
Semantic occupancy prediction jointly estimates the geometry and semantic categories of 3D space.
Early semantic scene completion methods infer volumetric structure from depth observations, while monocular approaches recover occupancy and semantics from RGB images~\cite{song2017sscnet,cao2022monoscene,yu2024iso}.
Camera-based methods develop image-to-volume lifting, sparse voxel queries, and structured spatial features~\cite{li2023voxformer,zhang2023occformer,li2023bridging,li2025sliceocc}.
In autonomous driving, surround-view and tri-perspective representations support scene-wide prediction~\cite{huang2023tpvformer,wei2023surroundocc}, with temporal modeling, self-supervision, and generative formulations extending occupancy to sequential scene understanding~\cite{ma2024cam4docc,chen2025casualocc,huang2024selfocc,zhang2025occnerf,li2026hierarchical,li2025scaling,li2025omninwm}.
OccAny further uses visual geometry models to predict urban occupancy from monocular, sequential, or surround-view images~\cite{cao2026occany}.

3D Gaussian Splatting~\cite{3dgs} motivates continuous occupancy representations built from spatial primitives with geometry and semantic attributes.
GaussianFormer~\cite{huang2024gaussianformer} represents scenes using sparse semantic Gaussians and obtains voxel predictions by aggregating nearby primitives through Gaussian-to-voxel splatting.
GaussianFormer-2~\cite{huang2024gaussianformer2} introduces probabilistic Gaussian superposition and distribution-based initialization to improve the allocation and aggregation of occupied-space primitives.
GaussianOcc~\cite{gan2024gaussianocc} exploits Gaussian splatting for self-supervised occupancy learning, while GaussianFormer3D~\cite{zhao2026gaussianformer3d} incorporates LiDAR geometry priors and camera features through 3D deformable attention.

For indoor scenes, SplatSSC~\cite{qian2026splatssc} combines depth-guided Gaussian initialization with decoupled geometric and semantic aggregation.
GPOcc~\cite{zhou2026gpocc} extends surface predictions along camera rays to construct volumetric Gaussian samples from visual geometry priors.
LegoOcc~\cite{zhou2026legoocc} couples occupancy geometry with language-aligned semantic embeddings through language-embedded Gaussians, enabling monocular open-vocabulary occupancy prediction.
In \methodname, Gaussian primitives encode occupied semantic evidence, with centers and covariances determined by predicted geometry and its uncertainty.
We pair these primitives with explicit free-space ray evidence and consolidate both in an incrementally constructed global memory.

\subsection{Online Occupancy Mapping and Spatial Memory}
Online scene understanding maintains spatial knowledge as observations arrive during exploration.
SCFusion~\cite{completescannet} combines incremental reconstruction with semantic scene completion from depth sequences.
EmbodiedOcc~\cite{wu2025embodiedocc} maintains an explicit global Gaussian memory, initializes Gaussians uniformly across the scene, and refines those inside the current camera frustum.
EmbodiedOcc++~\cite{wang2025embodiedocc_pp} introduces plane regularization and uncertainty-based selection of Gaussians for refinement.
RoboOcc~\cite{zhang2025roboocc} improves Gaussian interactions through opacity-guided and geometry-aware encoders, while SGR-OCC~\cite{guo2026sgrocc} combines uncertainty-aware feature lifting, ray-constrained refinement, and progressive training for temporal fusion.
Humanoid Occupancy~\cite{cui2025humanoidocc} extends embodied occupancy perception to panoramic multimodal sensing on humanoid platforms.

Incremental fusion provides another route to persistent occupancy mapping.
GPOcc~\cite{zhou2026gpocc} fuses per-frame Gaussians into a global representation through a training-free update strategy.
FreeOcc~\cite{jiang2026freeocc} integrates SLAM, incremental Gaussian mapping, and semantic features from pretrained vision-language models to construct open-vocabulary occupancy maps from monocular or RGB-D sequences.
These methods connect frame-level geometric predictions with global maps through spatial association and temporal integration.

\methodname develops a global semantic occupancy memory with on-demand allocation and explicit occupied and free-space evidence.
Local geometric predictions determine the position and spatial support of incoming semantic Gaussians, which are associated with existing memory elements or inserted as new elements.
Confidence-weighted fusion consolidates supporting observations, while free-space rays suppress contradicted occupancy and visibility-aware updates preserve occluded structure.
Spatial submaps and a building-level connectivity graph organize the accumulated evidence for continued exploration and occupancy queries.

\subsection{Visual Geometry Models as Observation Evidence}
Streaming visual geometry models also accumulate information through persistent reconstruction states.
Image-based geometry estimators provide depth and pointmap predictions from visual observations~\cite{Wang2024dust3r,leroy2024mast3r,wang2025vggt,keetha2026mapanything}.
For streaming reconstruction, CUT3R~\cite{wang2025cut3r} recurrently updates a latent state, Point3R~\cite{wu2025point3r} associates an explicit pointer memory with 3D locations, and TTT3R~\cite{chen2026ttt3r} adapts state updates according to the alignment between historical memory and incoming observations.
Their geometric outputs provide surface positions and temporal context for constructing spatial maps.

Geometric predictions and reconstruction states provide complementary information for evidence construction.
Predicted surfaces determine where an observation supports occupied structure, and the viewing rays identify observed free space before each surface hit.
In \methodname, surface geometry and uncertainty determine the centers and covariances of incoming semantic Gaussians, while image features support semantic and opacity prediction.
Causal fusion accumulates this evidence in memory elements that retain occupancy log-odds, semantic estimates, evidence weights, and observation counts.
This connects visual geometry estimation with occupancy memory that can be updated and queried throughout exploration.

\subsection{Benchmarks for Embodied Semantic Occupancy}
Semantic occupancy benchmarks define the observations, spatial targets, and evaluation protocols used to measure 3D scene understanding.
NYUv2, ScanNet, ScanNet++, and Matterport3D provide RGB-D observations and semantic scene annotations across different indoor capture settings~\cite{silberman2012nyuv2,scannet,scannetpp,Matterport3D2017}.
ScanNet provides posed perspective frames, reconstructed surfaces, and semantic annotations that support local and room-level occupancy evaluation.
ScanNet++ supplies higher-fidelity reconstructions and visual observations, while Matterport3D supplies panoramic observations across connected rooms and floors.
These capture settings support evaluation of both local scene prediction and memory construction during exploration.
In autonomous driving, OpenOccupancy, Occ3D, SSCBench, and OpenScene establish large-scale evaluation of surrounding occupancy and semantic scene completion~\cite{wang2023openoccupancy,tian2023occ3d,li2024sscbench,openscene2023}.

Indoor occupancy benchmarks develop complementary prediction and mapping settings.
Occ-ScanNet evaluates local semantic occupancy from monocular observations, while EmbodiedOcc-ScanNet organizes posed observation sequences for room-level online occupancy prediction~\cite{yu2024iso,wu2025embodiedocc}.
CompleteScanNet supplies completed scene geometry for semantic scene completion~\cite{completescannet}.
EmbodiedScan broadens indoor evaluation to multimodal egocentric perception with dense semantic occupancy annotations~\cite{wang2024embodiedscan}.
Humanoid Occupancy introduces panoramic multimodal occupancy data for humanoid robots~\cite{cui2025humanoidocc}, and ReplicaOcc, introduced with FreeOcc~\cite{jiang2026freeocc}, supports indoor open-vocabulary occupancy evaluation.

\datasetname establishes a unified benchmark for evaluating semantic occupancy memory from local observations to connected buildings.
Its three tracks cover local prediction, causal room-level mapping, and building-level exploration, with shared semantic labels and occupancy-state definitions.
View-conditioned annotations and validity masks support consistent evaluation across perspective frames and pano-centric observation groups.
Semantic mIoU is computed over the 11 occupied classes inherited from Occ-ScanNet, while free space is evaluated as a separate occupancy state.
This common framework connects occupancy accuracy with revisit consistency, memory footprint, and query latency, enabling assessment of how accumulated observations support reliable and scalable spatial memory.